\documentclass[runningheads]{llncs}

\usepackage[T1]{fontenc}
\usepackage[utf8]{inputenc}
\usepackage{microtype}
\usepackage{graphicx}
\usepackage{booktabs}
\usepackage{multirow}
\usepackage{array}
\usepackage{amsmath,amssymb,mathtools,bm}
\usepackage{xcolor}
\usepackage{enumitem}
\usepackage{placeins}
\usepackage{url}
\usepackage{hyperref}
\usepackage{marvosym}
\usepackage{orcidlink}
\usepackage[nameinlink,capitalize]{cleveref}

\hypersetup{
  colorlinks=true,
  linkcolor=blue!55!black,
  citecolor=blue!55!black,
  urlcolor=blue!55!black,
  pdfauthor={Zonglin Yang, Huilan Ma, Xudan Zheng, Yuejun Xie},
  pdftitle={UOT-Gap: A Variational Principle for the Modality Gap in Vision-Language Models}
}

\newcommand{\papername}{\textsc{UOT-Gap}}
\newcommand{\correspondingemail}[1]{\textsuperscript{\href{mailto:#1}{\textcolor{black}{\Letter}}}}
\newcommand{\R}{\mathbb{R}}
\newcommand{\E}{\mathbb{E}}
\newcommand{\KL}{\mathrm{KL}}

\newcommand{\one}{\mathbf{1}}
\newcommand{\Sphere}{\mathbb{S}^{d-1}}
\newcommand{\ip}[2]{\left\langle #1,#2 \right\rangle}
\newcommand{\norm}[1]{\left\lVert #1 \right\rVert}
\newcommand{\argmin}{\mathop{\mathrm{argmin}}}

\newcommand{\OT}{\mathsf{OT}}
\newcommand{\Gap}{\mathsf{Gap}}
\newcommand{\Cost}{\mathsf{Cost}}
\newcommand{\Marg}{\mathsf{Marg}}
\newcommand{\Coupl}{\mathsf{Coupl}}
\newcommand{\dd}{\mathrm{d}}

\spnewtheorem{assumption}[theorem]{Assumption}{\bfseries}{\itshape}
\crefname{theorem}{Theorem}{Theorems}
\crefname{assumption}{Assumption}{Assumptions}
\crefname{figure}{Fig.}{Figs.}
\crefname{table}{Table}{Tables}
\crefname{equation}{Eq.}{Eqs.}

\title{\papername: A Variational Principle for the Modality Gap in Vision--Language Models via Unbalanced Optimal Transport}
\titlerunning{\papername}

\author{Zonglin Yang\,\orcidlink{0009-0000-5732-5699}\correspondingemail{3258244847@qq.com} \and Huilan Ma \and Xudan Zheng \and Yuejun Xie}
\authorrunning{Z. Yang et al.}
\institute{Guangdong Police College\\
\email{3258244847@qq.com}\\
\email{1451727521@qq.com}\\
\email{3113434702@qq.com}\\
\email{18320142522@163.com}}

\begin{document}
\maketitle

\begin{abstract}
Vision--language models such as CLIP embed images and text in a shared space, where modality-specific distributions often remain separated. Existing accounts connect this \emph{modality gap} to initialization, contrastive dynamics, and information imbalance, while its distributional and pairwise contributions to retrieval remain unresolved. We introduce \papername, a training-free variational diagnostic that models frozen image and text embeddings with unbalanced entropic optimal transport (UOT). The UOT optimum separates transport, coupling complexity, and marginal mass variation; a complementary \emph{pair-aware} residual compares observed image--caption pairs with the UOT soft matching. On Flickr8K and COCO-1K with frozen CLIP, OpenCLIP, and SigLIP encoders, caption degradation reduces Flickr8K Recall@1 from 0.559 to 0.003. Across six dataset--model conditions, the pair-aware residual tracks retrieval degradation with mean absolute Spearman 0.973, compared with 0.392 for the mean gap. The association remains stable across five random COCO-1K subsets at $0.954\pm0.026$, with a minimum of 0.943. UOT barycentric updates reduce the transport objective while degrading retrieval, distinguishing geometric objective descent from task improvement. These results establish \papername{} as a diagnostic for caption quality, modality alignment, and retrieval robustness.
\keywords{Vision-language models \and Modality gap \and Unbalanced optimal transport \and Cross-modal retrieval}
\end{abstract}

\section{Introduction}

Contrastive vision--language models (VLMs), such as CLIP \cite{radford2021learning} and SigLIP \cite{zhai2023sigmoid}, map images and captions into a shared representation space. Their embeddings often form modality-specific clouds, producing the \emph{modality gap} \cite{liang2022mind}. Previous studies connect this structure to initialization and contrastive optimization \cite{liang2022mind}, gradient-flow dynamics and mismatched pairs \cite{yaras2024explaining}, and information imbalance between images and captions \cite{schrodi2025two}. A remaining diagnostic challenge is to identify a variational quantity that separates distributional mismatch from broken image--caption correspondence.

We address this challenge with \emph{unbalanced entropic optimal transport} (UOT). Balanced optimal transport enforces complete mass conservation, whereas VLM representations contain modality-specific factors: images may include visual content omitted by captions, and captions may express abstractions without localized visual counterparts. UOT assigns a penalty to mass deletion or creation and thereby represents this information asymmetry directly.

Given frozen image embeddings $X=\{x_i\}_{i=1}^n$ and text embeddings $Y=\{y_j\}_{j=1}^m$ on the unit sphere, empirical weights $a,b$, and cosine cost $C_{ij}=1-\ip{x_i}{y_j}$, we solve
\begin{align}
\pi^\star=\argmin_{\pi\ge0}\;
\langle C,\pi\rangle
+\varepsilon\KL(\pi\|ab^\top)
+\rho_x\KL(\pi\one\|a)
+\rho_y\KL(\pi^\top\one\|b).
\label{eq:uot-intro}
\end{align}
The optimized objective yields transport, coupling, and marginal components. For paired retrieval datasets, we further define a pair-aware residual that compares the observed diagonal pairing cost with the UOT matching cost. A random caption permutation isolates the resulting distinction by preserving the text marginal distribution while disrupting pairwise retrieval.

Using only frozen embeddings, \papername{} can screen caption quality, monitor modality alignment during training, and audit retrieval robustness without encoder fine-tuning.

\begin{figure}[t]
\centering
\includegraphics[width=\linewidth]{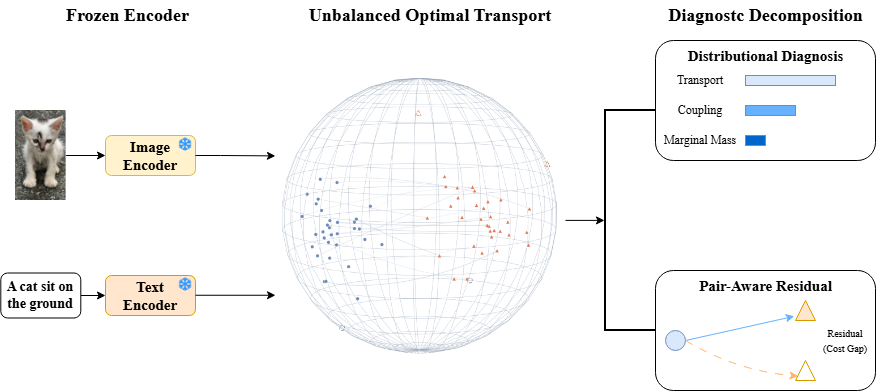}
\caption{Overview. \papername{} interprets the modality gap as the residual of an unbalanced transport problem. Distributional components diagnose marginal and coupling mismatch; the pair-aware residual diagnoses whether observed image--caption pairs are worse than the best soft UOT matching.}
\label{fig:overview}
\end{figure}

Our contributions are:
\begin{enumerate}[leftmargin=1.4em,itemsep=0.1em,topsep=0.2em]
\item We formulate the VLM modality gap as the residual of an unbalanced entropic optimal transport problem between frozen image and text embeddings, and introduce a pair-aware UOT residual that separates marginal distribution mismatch from broken image--caption alignment.
\item We prove three results that justify the diagnostic: transport bounds the classical mean gap, missing modality-specific mass induces an unavoidable UOT cost or marginal penalty, and UOT barycenters provide a first-order descent direction for the UOT objective.
\item We validate the diagnostic across Flickr8K, COCO-1K, synthetic controlled data, three frozen encoders, five random COCO subsets, caption degradation, hyperparameter sweeps, and post-hoc correction baselines, showing that the pair-aware residual tracks retrieval degradation more reliably than the classical mean gap.
\end{enumerate}

\section{Related Work}

\paragraph{Modality gap in VLMs.}
CLIP learns transferable visual representations by aligning images and text with a contrastive loss \cite{radford2021learning}. Liang et al. \cite{liang2022mind} showed that modalities can remain separated in the shared embedding space and connected this gap to initialization and optimization. Yaras et al. \cite{yaras2024explaining} analyze gradient-flow dynamics and identify mismatched pairs and temperature as mechanisms that preserve the gap. Schrodi et al. \cite{schrodi2025two} link information imbalance to both modality gap and object bias. These causal accounts motivate our complementary variational diagnostic, whose residuals quantify the cost and mass variation required for cross-modal matching.

\paragraph{Vision--language pretraining and retrieval.}
Vision--language pretraining has evolved from cross-modal Transformer encoders \cite{lu2019vilbert,chen2020uniter} to large-scale dual encoders and frozen-backbone image--text models \cite{jia2021scaling,li2022blip,li2023blip}. These models improve retrieval and transfer through greater alignment capacity. \papername{} complements this progress with a post-hoc diagnostic that distinguishes marginal distribution shift from broken pair identity across frozen encoders.

\paragraph{Optimal transport and unbalanced matching.}
Optimal transport provides a geometry for comparing distributions \cite{villani2009optimal,peyre2019computational}. Entropic regularization enables fast Sinkhorn scaling \cite{cuturi2013sinkhorn}, and unbalanced optimal transport relaxes exact marginal constraints with divergence penalties \cite{chizat2018scaling,pham2020unbalanced,janati2020entropic}. Sinkhorn divergences and their sample behavior motivate practical regularized distribution comparison \cite{feydy2019interpolating,genevay2019sample}. Practical solvers are available in the Python Optimal Transport library \cite{flamary2021pot}. We treat the optimized UOT residual as a diagnostic signal that exposes the structure of relaxed cross-modal matching.

\paragraph{Multimodal identifiability and information asymmetry.}
Multimodal contrastive learning may identify shared latent factors while leaving modality-specific factors unresolved \cite{daunhawer2023identifiability}. This view motivates our missing-attribute analysis: private factors make relaxed mass matching the appropriate object, allowing UOT to separate alignable shared mass from residual modality-specific mass.

\section{Method}

\subsection{Embeddings and UOT Objective}

Let $f_\theta$ be an image encoder and $g_\phi$ a text encoder. For image--caption pairs $\{(I_i,T_i)\}_{i=1}^n$, define normalized embeddings
\begin{align}
x_i=\frac{f_\theta(I_i)}{\norm{f_\theta(I_i)}_2},\qquad
y_i=\frac{g_\phi(T_i)}{\norm{g_\phi(T_i)}_2},\qquad
x_i,y_i\in\Sphere .
\end{align}
The empirical measures are $\hat\mu=\sum_i a_i\delta_{x_i}$ and $\hat\nu=\sum_j b_j\delta_{y_j}$, typically with uniform weights. The classical mean gap is
\begin{align}
\Gap_{\rm mean}(\hat\mu,\hat\nu)
=\norm{\sum_i a_ix_i-\sum_j b_jy_j}_2.
\end{align}
The mean gap captures global centroid displacement; higher-order geometry, neighborhood structure, and pair identity remain outside this first-moment statistic.

For a nonnegative coupling $\pi\in\R_+^{n\times m}$, denote row and column marginals by $r(\pi)=\pi\one_m$ and $c(\pi)=\pi^\top\one_n$. With generalized KL divergence, the UOT objective is
\begin{align}
\mathcal{J}_{\varepsilon,\rho_x,\rho_y}(\pi;X,Y)
=\langle C,\pi\rangle
+\varepsilon\KL(\pi\|ab^\top)
+\rho_x\KL(r(\pi)\|a)
+\rho_y\KL(c(\pi)\|b).
\label{eq:uot-objective}
\end{align}
Let $\pi^\star$ minimize \cref{eq:uot-objective}. We define
\begin{align}
\Cost&=\langle C,\pi^\star\rangle, &
\Coupl&=\KL(\pi^\star\|ab^\top), &
\Marg&=\KL(r(\pi^\star)\|a)+\KL(c(\pi^\star)\|b).
\end{align}
These terms form a distributional decomposition: $\Cost$ measures cross-modal matching cost, $\Coupl$ measures complexity relative to independent matching, and $\Marg$ measures mass variation through deletion or creation.

\subsection{Pair-Aware UOT Residual}

Retrieval depends on both the marginal distributions and the observed image--caption pairing. A random caption permutation preserves the text distribution while collapsing retrieval. We therefore report the paired cost
\begin{align}
\Cost_{\rm pair}(X,Y)=\frac{1}{n}\sum_{i=1}^{n}C_{ii},
\end{align}
and the pair-aware UOT residual
\begin{align}
\Gap_{\rm pair}(X,Y)
=\Cost_{\rm pair}(X,Y)-\frac{\Cost(X,Y)}{\sum_{ij}\pi^\star_{ij}}.
\label{eq:pair-uot-gap}
\end{align}
Intuitively, $\Gap_{\rm pair}$ is large when the observed pairs are much worse than the best soft cross-modal matching implied by UOT. The reported score is
\begin{align}
\papername_{\varepsilon,\rho}(X,Y)
=\left[\Gap_{\rm pair}(X,Y)+\Marg(X,Y)+0.1\,\Coupl(X,Y)\right]_+^{1/2}.
\label{eq:uotgap-score}
\end{align}
For distributional diagnosis, the components should be inspected separately. For retrieval degradation diagnosis, $\Gap_{\rm pair}$ is the primary UOT-derived statistic because it explicitly evaluates the observed image--caption pair.

\subsection{Barycentric Correction}

The UOT plan also induces a lightweight post-hoc correction. For rows with positive transported mass
$r_i^\star=\sum_j\pi_{ij}^\star$, define the text barycenter
\begin{align}
b_i=\frac{\sum_j\pi_{ij}^\star y_j}{r_i^\star}.
\end{align}
Because CLIP-like embeddings are normalized, we move along the tangent direction on the sphere:
\begin{align}
v_i=P_{x_i}^{\perp}b_i=b_i-\ip{x_i}{b_i}x_i,\qquad
x_i'=\frac{x_i+\alpha v_i}{\norm{x_i+\alpha v_i}_2}.
\label{eq:barycentric-correction}
\end{align}
This update serves as a diagnostic intervention that tests whether the UOT matching geometry reduces transport residuals and modality separability. Because its direction follows soft matching instead of observed diagonal pairs, we evaluate retrieval jointly with every gap metric.

\section{Theory}

We state three results that motivate the diagnostic. Full derivations use Jensen's inequality, Pinsker's inequality, pointwise minimization over retained private mass, and the envelope theorem for the optimized UOT value.

\begin{theorem}[Balanced transport certificate]
\label{thm:balanced-gap}
Let $\mu,\nu$ be probability measures on $\Sphere$, and let $c(x,y)=1-\ip{x}{y}$. For any coupling $\gamma\in\Pi(\mu,\nu)$,
\begin{align}
\norm{m_\mu-m_\nu}_2^2\le2\int c(x,y)\,\dd\gamma(x,y).
\end{align}
Consequently $\Gap_{\rm mean}^2(\mu,\nu)\le 2\OT_c(\mu,\nu)$.
\end{theorem}

The theorem provides a transport certificate for the mean gap: low cosine transport cost implies small centroid separation. Its scope is global displacement; local mismatch and missing modality-specific factors require the relaxed decomposition developed below.

\paragraph{Proof sketch.}
Let $(X,Y)\sim\gamma$. Since $\gamma$ has marginals $\mu$ and $\nu$,
$\E[X-Y]=m_\mu-m_\nu$. Jensen's inequality gives
$\norm{m_\mu-m_\nu}_2^2\le\E\norm{X-Y}_2^2$. On the unit sphere,
$\norm{x-y}_2^2=2-2\ip{x}{y}=2c(x,y)$. Taking the infimum over couplings yields the OT bound.

\begin{theorem}[UOT certificate for the mean gap]
\label{thm:uot-certificate}
Let $\mu,\nu$ be probability measures on $\Sphere$. Let $\pi$ be any nonzero finite coupling whose normalized marginals $\bar r,\bar c$ are absolutely continuous with respect to $\mu,\nu$. Then
\begin{align}
\Gap_{\rm mean}(\mu,\nu)
\le
\sqrt{2\int c(x,y)\,\dd\bar\pi(x,y)}
+\sqrt{2\KL(\bar r\|\mu)}
+\sqrt{2\KL(\bar c\|\nu)} .
\label{eq:uot-certificate-bound}
\end{align}
\end{theorem}

\paragraph{Proof sketch.}
Decompose
$m_\mu-m_\nu=(m_\mu-m_{\bar r})+(m_{\bar r}-m_{\bar c})+(m_{\bar c}-m_\nu)$.
The middle term is controlled by \cref{thm:balanced-gap} applied to the normalized coupling $\bar\pi$. The two marginal terms are bounded by total variation between the normalized marginals and the original measures, and Pinsker's inequality converts those total-variation terms into KL terms. Transport and marginal deviations therefore jointly certify the classical mean gap and ground the UOT residuals in a measurable geometric quantity.

\begin{assumption}[Separated image-private mass]
\label{ass:private-mass}
There exists $A\subset\Sphere$ with $\mu(A)=\eta>0$ such that $c(x,y)\ge\delta>0$ for all $x\in A$ and $y\in{\rm supp}(\nu)$. We interpret $A$ as image-private mass absent or underspecified in captions.
\end{assumption}

\begin{theorem}[Missing-attribute lower bound]
\label{thm:missing-attribute}
Under \cref{ass:private-mass}, every finite coupling $\pi$ for the semi-unbalanced objective
$\mathcal{J}_{\rho}^{x}(\pi)=\int c\,\dd\pi+\rho\KL(\pi_X\|\mu)$ satisfies
\begin{align}
\mathcal{J}_{\rho}^{x}(\pi)
\ge
\rho\eta\left(1-e^{-\delta/\rho}\right).
\label{eq:missing-attribute-lower-bound}
\end{align}
If the private attribute is uniformly distributed over $k$ states and only one state is text-alignable, then $\eta=1-1/k=1-e^{-H(U)}$.
\end{theorem}

Private visual mass therefore contributes a positive UOT residual through either transport cost or marginal KL.

\paragraph{Proof sketch.}
On the private set $A$, let $s(x)\in[0,1]$ denote the fraction of image mass retained by a coupling. Transporting retained mass costs at least $\delta s(x)$, while deleting mass pays $\rho(s(x)\log s(x)-s(x)+1)$ under generalized KL. Minimizing the pointwise function over $s$ gives $s^\star=e^{-\delta/\rho}$ and value $\rho(1-e^{-\delta/\rho})$. Integrating over $\mu(A)=\eta$ proves the lower bound. The entropy expression follows by setting $\eta=1-1/k$ for a uniformly distributed private factor with one alignable state.

\begin{theorem}[First-order UOT correction descent]
\label{thm:correction-descent}
For each image embedding define $r_i^\star=\sum_j\pi^\star_{ij}$ and
$b_i=(\sum_j\pi^\star_{ij}y_j)/r_i^\star$. Update
\begin{align}
x_i(\alpha)=
\frac{x_i+\alpha P_{x_i}^{\perp}b_i}
{\norm{x_i+\alpha P_{x_i}^{\perp}b_i}_2}.
\end{align}
Let $\Phi(X,Y)=\min_{\pi\ge0}\mathcal{J}_{\varepsilon,\rho_x,\rho_y}(\pi;X,Y)$.
For sufficiently small $\alpha>0$,
\begin{align}
\Phi(X(\alpha),Y)
\le
\Phi(X,Y)
-\alpha\sum_{i:r_i^\star>0} r_i^\star\norm{P_{x_i}^{\perp}b_i}_2^2
+O(\alpha^2).
\end{align}
\end{theorem}

The guarantee concerns first-order descent of the UOT objective; retrieval response is evaluated empirically.

\paragraph{Proof sketch.}
For fixed $\pi^\star$, only the cosine cost term changes to first order under the update. The derivative of
$1-\ip{x_i(\alpha)}{y_j}$ at $\alpha=0$ equals $-\ip{P_{x_i}^{\perp}b_i}{y_j}$. Summing with weights $\pi_{ij}^\star$ yields
$-r_i^\star\norm{P_{x_i}^{\perp}b_i}_2^2$ for row $i$. The envelope theorem transfers this fixed-plan descent to the optimized value up to $O(\alpha^2)$ terms, assuming local stability of the entropic UOT optimum.

\section{Experiments}
\label{sec:experiments}

\subsection{Setup}

We evaluate frozen CLIP ViT-B/32 \cite{radford2021learning}, OpenCLIP ViT-B/32 \cite{ilharco2021openclip}, and SigLIP B/16 \cite{zhai2023sigmoid} encoders. Flickr8K \cite{hodosh2013framing} and the COCO Karpathy split \cite{lin2014microsoft,karpathy2015deep} are evaluated with $n=1000$ image--caption pairs, and five independent COCO subsets measure sampling stability. Caption conditions include full, half, quarter, noun-only, attribute-dropped, and random captions. The random condition cyclically shifts captions, preserving the text marginal distribution while breaking pair identity.

Unless otherwise stated, we use cosine cost, $\varepsilon=0.08$, $\rho=0.8$, generalized Sinkhorn iterations with threshold $10^{-7}$, and report Recall@1/5/10, modality-probe AUC, logit entropy, $\Gap_{\rm mean}$, $\Cost_{\rm pair}$, $\Gap_{\rm pair}$, and \papername{}. We compute metrics from cached frozen embeddings; no encoder is fine-tuned. The modality probe is a logistic classifier trained to distinguish image from text embeddings. The synthetic experiment samples normalized Gaussian sphere distributions with controlled mean separation and concentration imbalance, allowing the marginal and transport terms to be tested independently of caption semantics.

\begin{figure}[t]
\centering
\includegraphics[width=\linewidth]{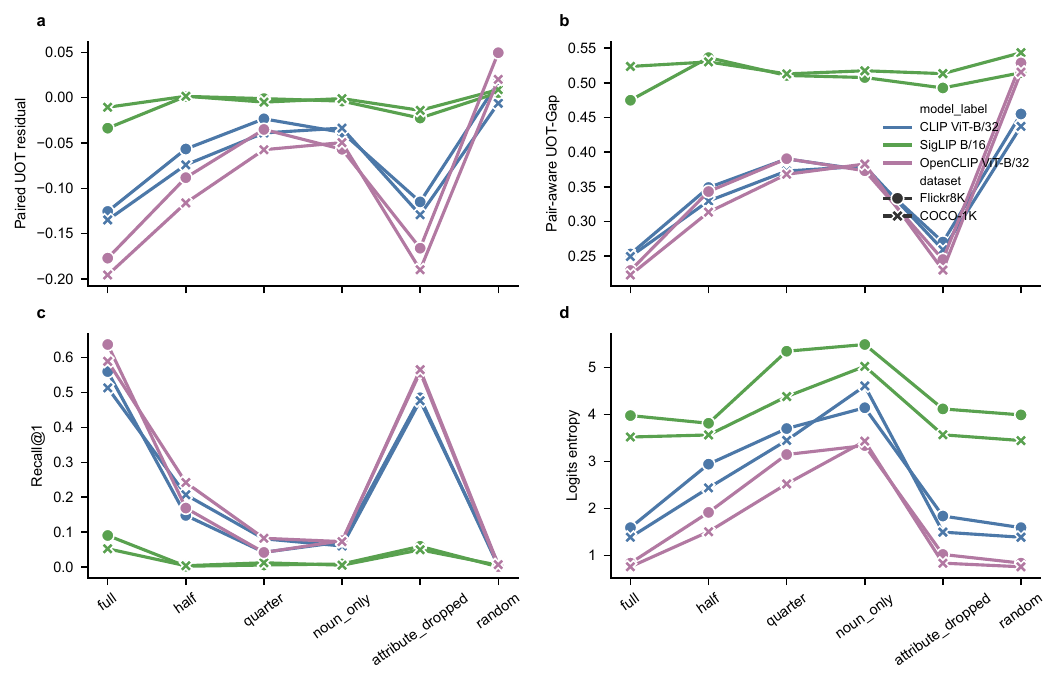}
\caption{Caption degradation on Flickr8K and COCO-1K. Pair-aware UOT residuals and the composite \papername{} score increase as observed caption pairs become less aligned, while Recall@1 decreases.}
\label{fig:caption-degradation}
\end{figure}

\subsection{Retrieval Degradation}

\begin{table}[t]
\centering
\caption{Caption degradation on Flickr8K with CLIP ViT-B/32. The random condition preserves the caption distribution while breaking observed pairing, isolating the pair-aware signal.}
\label{tab:caption-degradation}
\setlength{\tabcolsep}{4.2pt}
\begin{tabular}{lcccccc}
\toprule
Caption & Mean & PairCost & Pair-UOT & \papername{} & R@1 & Entropy \\
\midrule
Full & 0.825 & 0.681 & -0.126 & 0.253 & 0.559 & 1.595 \\
Half & 0.922 & 0.739 & -0.057 & 0.349 & 0.148 & 2.945 \\
Quarter & 0.981 & 0.773 & -0.023 & 0.390 & 0.043 & 3.701 \\
Noun-only & 0.970 & 0.761 & -0.038 & 0.374 & 0.072 & 4.143 \\
Attribute-drop & 0.863 & 0.691 & -0.115 & 0.270 & 0.484 & 1.842 \\
Random & 0.825 & 0.824 & 0.017 & 0.455 & 0.003 & 1.595 \\
\bottomrule
\end{tabular}
\end{table}

\begin{table}[t]
\centering
\caption{Spearman correlation between gap metrics and retrieval degradation across caption conditions. Higher absolute values indicate stronger monotone association with Recall@1 drop.}
\label{tab:correlation}
\setlength{\tabcolsep}{3.7pt}
\begin{tabular}{llccccc}
\toprule
Dataset & Model & Mean & Paired & Pair-UOT & \papername{} & Entropy \\
\midrule
COCO-1K & CLIP & 0.429 & 1.000 & 1.000 & 1.000 & 0.290 \\
COCO-1K & OpenCLIP & 0.143 & 1.000 & 1.000 & 1.000 & 0.290 \\
COCO-1K & SigLIP & 0.841 & 0.464 & 0.841 & 0.464 & 0.058 \\
Flickr8K & CLIP & 0.143 & 1.000 & 1.000 & 1.000 & 0.232 \\
Flickr8K & OpenCLIP & 0.429 & 1.000 & 1.000 & 1.000 & 0.371 \\
Flickr8K & SigLIP & 0.371 & 0.657 & 1.000 & 0.943 & -0.143 \\
\midrule
Mean absolute & -- & 0.392 & 0.853 & 0.973 & 0.901 & 0.231 \\
\bottomrule
\end{tabular}
\end{table}

The main result is that pair-aware UOT statistics track retrieval degradation more reliably than the classical mean gap in this benchmark. On Flickr8K with CLIP ViT-B/32, Recall@1 drops from 0.559 for full captions to 0.148 for half captions, 0.043 for quarter captions, and 0.003 for random captions. The mean gap is unchanged by random permutation because the caption distribution is unchanged, while $\Cost_{\rm pair}$ increases from 0.681 to 0.824 and $\Gap_{\rm pair}$ increases from $-0.126$ to 0.017. Across all fixed-parameter dataset--model conditions, $\Gap_{\rm pair}$ reaches best absolute Spearman 1.000 and mean absolute Spearman 0.973, compared with 0.841 best and 0.392 mean for $\Gap_{\rm mean}$.

The COCO subset analysis shows stable behavior across five independent 1K samples with CLIP ViT-B/32. Full-caption Recall@1 is $0.514\pm0.009$, quarter-caption Recall@1 is $0.078\pm0.005$, and random-caption Recall@1 is $0.001\pm0.001$. Across these seeds, $\Gap_{\rm pair}$ reaches mean Spearman $0.954\pm0.026$ with a minimum of 0.943, whereas $\Gap_{\rm mean}$ averages $0.269\pm0.142$. The recurring association supports frozen-embedding subsets as a stable low-compute diagnostic setting.

\begin{figure}[t]
\centering
\includegraphics[width=\linewidth]{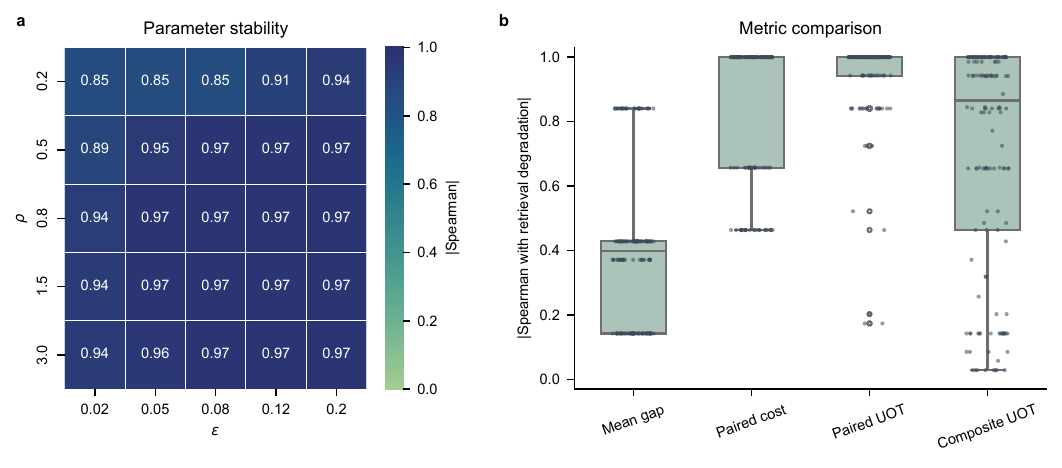}
\caption{UOT hyperparameter sensitivity. Across 150 parameter--dataset--model combinations, the pair-aware UOT residual has mean absolute Spearman 0.947 and median 1.000; 142/150 combinations reach $|\rho_s|\ge0.8$.}
\label{fig:param-sweep}
\end{figure}

\subsection{Synthetic Data and Hyperparameter Sensitivity}

\begin{figure}[t]
\centering
\includegraphics[width=\linewidth]{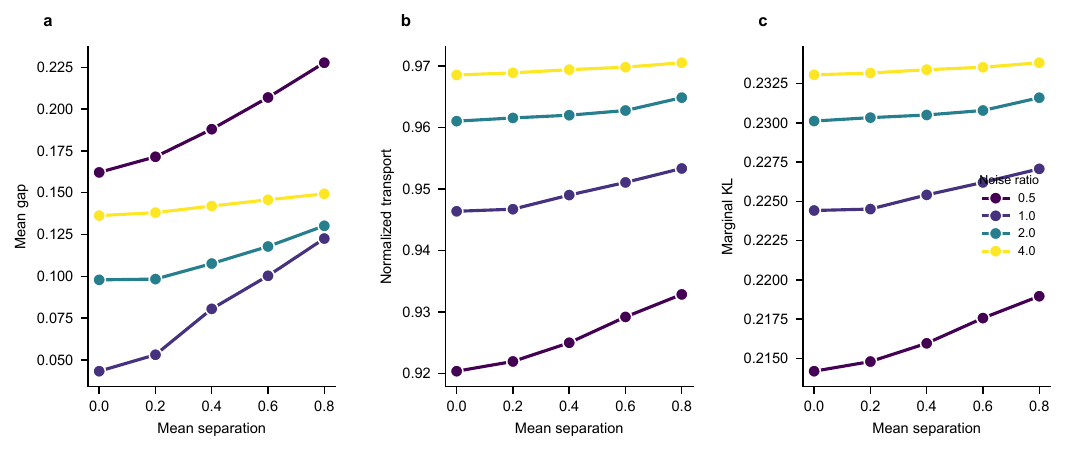}
\caption{Synthetic controlled data. Mean separation and concentration imbalance are varied on normalized Gaussian sphere samples. UOT transport and marginal terms respond to distributional shifts, while the mean gap isolates global displacement.}
\label{fig:synthetic}
\end{figure}

On normalized Gaussian sphere samples, mean separation increases the mean gap, whereas concentration imbalance increases UOT marginal and transport terms. The mean gap therefore isolates global displacement, while UOT components respond to higher-order distributional mismatch. Changes in concentration can increase the UOT terms even when centroid separation remains modest.

We also sweep $\varepsilon\in\{0.02,0.05,0.08,0.12,0.20\}$ and $\rho\in\{0.2,0.5,0.8,1.5,3.0\}$. Across 150 parameter--dataset--model combinations, the pair-aware residual has mean absolute Spearman 0.947 and median 1.000 with retrieval degradation; 142 combinations reach $|\rho_s|\ge0.8$. The pair-aware association is stable over this practical grid, while the raw decomposition values vary with $\varepsilon$ and $\rho$ and are reported together with both parameters.

\subsection{Model Sweep}

\begin{figure}[t]
\centering
\includegraphics[width=\linewidth]{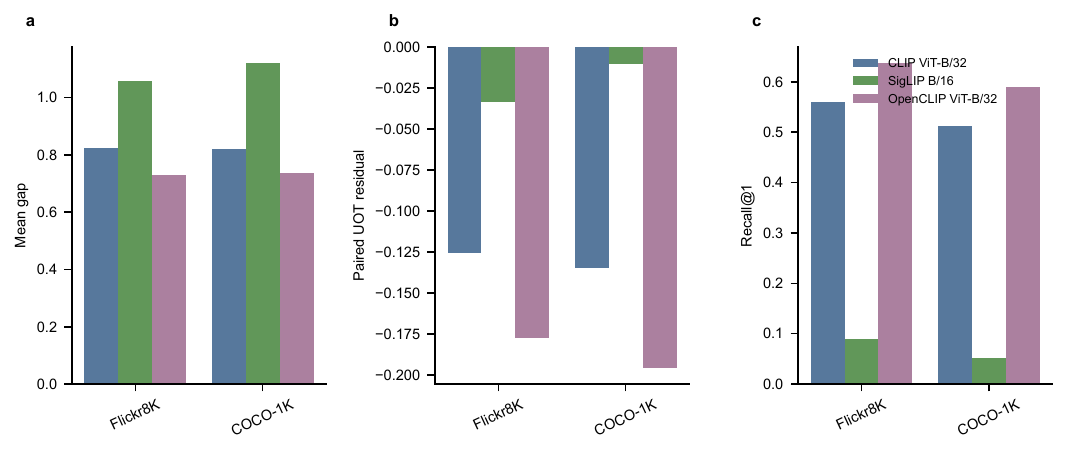}
\caption{Model sweep over frozen CLIP, OpenCLIP, and SigLIP checkpoints. Pair-aware UOT residuals are most informative for CLIP/OpenCLIP retrieval degradation settings and remain model-dependent for SigLIP.}
\label{fig:model-sweep}
\end{figure}

\cref{fig:model-sweep} demonstrates diagnostic behavior across three encoder families. CLIP and OpenCLIP show the clearest monotone relationship between pair-aware UOT residuals and caption degradation on Flickr8K and COCO-1K. SigLIP is more model-dependent: on COCO-1K, the mean gap has high correlation in one setting, while the composite \papername{} score is weaker because it combines marginal and coupling terms with the pair residual. This model dependence motivates reporting both the composite score and the raw pair-aware residual.

\subsection{Post-hoc Correction}

We compare raw embeddings with mean-shift removal, orthogonal Procrustes alignment, and UOT barycentric correction. Procrustes is a strong pair-preserving baseline based on global orthogonal alignment.

\begin{table}[t]
\centering
\caption{Post-hoc correction on Flickr8K with CLIP ViT-B/32. Lower AUC means less linear modality separability; higher retrieval is better.}
\label{tab:correction}
\setlength{\tabcolsep}{3.8pt}
\begin{tabular}{lcccccc}
\toprule
Method & Mean$\downarrow$ & PairCost$\downarrow$ & Pair-UOT & AUC$\downarrow$ & R@1$\uparrow$ & R@5$\uparrow$ \\
\midrule
Raw & 0.825 & 0.681 & -0.126 & 1.000 & 0.559 & 0.824 \\
Mean-shift & 0.009 & 0.341 & -0.121 & 0.270 & 0.356 & 0.609 \\
Procrustes & 0.014 & 0.195 & -0.219 & 0.317 & 0.829 & 0.947 \\
UOT $\alpha=0.2$ & 0.713 & 0.574 & -0.123 & 1.000 & 0.521 & 0.805 \\
UOT $\alpha=0.4$ & 0.570 & 0.447 & -0.114 & 1.000 & 0.430 & 0.708 \\
UOT $\alpha=0.6$ & 0.419 & 0.324 & -0.094 & 1.000 & 0.349 & 0.572 \\
\bottomrule
\end{tabular}
\end{table}

\begin{figure}[!t]
\centering
\includegraphics[width=0.92\linewidth]{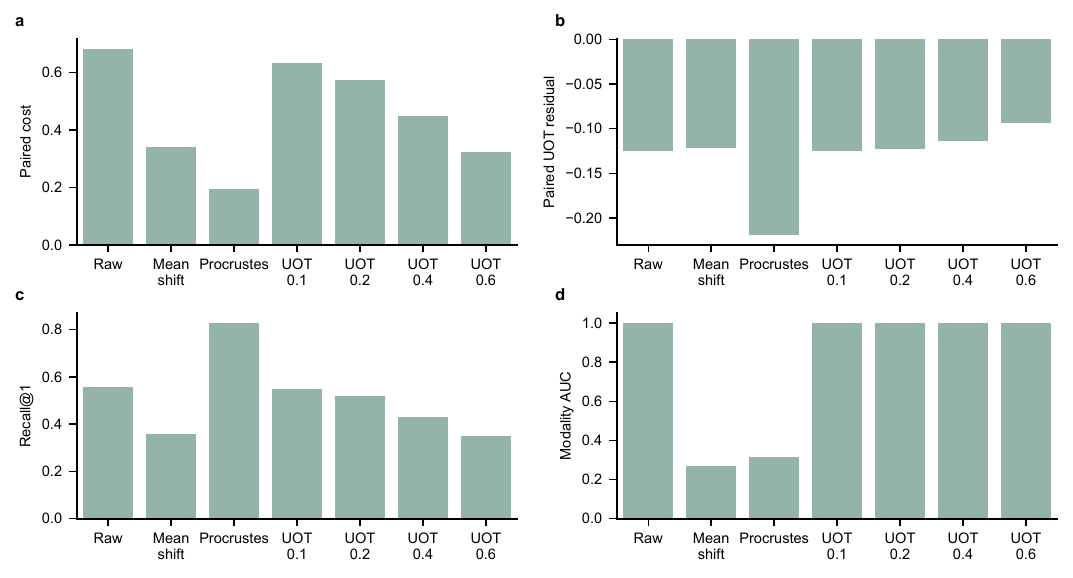}
\caption{Correction trade-off. UOT barycentric steps reduce paired cost and UOT residual as retrieval declines, while Procrustes provides the strongest retrieval correction in this frozen-embedding run.}
\label{fig:correction}
\end{figure}

UOT barycentric correction follows \cref{thm:correction-descent}: it decreases UOT-related objective terms, with paired cost dropping from 0.681 for raw CLIP embeddings to 0.324 at $\alpha=0.6$. Retrieval simultaneously decreases from R@1 0.559 to 0.349, whereas Procrustes increases R@1 to 0.829. This result separates objective descent from task performance and supports barycentric correction as a diagnostic intervention.

\FloatBarrier

\section{Discussion and Limitations}

The central finding is that the modality gap has two diagnostic levels. Distributional UOT components quantify transport, coupling complexity, and marginal mass variation, while the pair-aware residual measures how observed image--caption pairs differ from the UOT soft matching. The random-caption control isolates this distinction: it preserves the caption distribution, collapses retrieval, and increases the pair-aware residual.

Across the tested datasets and encoders, the pair-aware residual tracks caption-induced retrieval degradation more consistently than the mean-gap baseline. Distributional UOT terms answer a complementary question and vary more across encoders. We therefore recommend component-level reporting: transport and marginal terms for distribution comparison, and the pair-aware residual for paired retrieval diagnosis.

The correction experiment further separates geometric optimization from downstream utility. UOT barycentric steps reduce paired cost and UOT objective terms, as predicted by the descent theorem, while retrieval declines as the diagonal pair structure is smoothed. Procrustes preserves paired supervision and performs better as a retrieval correction. This comparison defines \papername{} as an analysis tool and its barycentric update as a diagnostic intervention.

The present scope covers frozen global embeddings and 1K retrieval subsets. Five independent COCO samples establish subset stability, while full COCO evaluation remains an important extension. The missing-attribute lower bound uses a stylized separation assumption, the pair-aware residual requires paired image--caption data, and UOT hyperparameters affect the decomposition values. Future evaluation should include larger retrieval benchmarks, additional VLM families, and distributional baselines such as MMD \cite{gretton2012kernel}, energy distance \cite{szekely2013energy}, and Sinkhorn divergence \cite{feydy2019interpolating}. Pair-aware comparisons with captioning metrics such as CLIPScore \cite{hessel2021clipscore} will further define the diagnostic range.

\subsection*{Reproducibility and Ethics}

The experiments use frozen forward passes and UOT/Sinkhorn solvers. The implementation stores embeddings, caption-degradation metadata, random seeds, UOT hyperparameter grids, and plotting scripts. Task metrics accompany every geometric correction because semantic errors can persist after distributional alignment.

\section{Conclusion}

We introduced \papername, a variational framework that interprets the VLM modality gap through unbalanced entropic optimal transport. The theory connects transport and marginal variation to the mean gap and missing modality-specific mass, and establishes a first-order descent direction for UOT barycentric updates. Empirically, distributional components decompose modality imbalance, while the pair-aware residual tracks retrieval degradation across caption conditions, encoder families, hyperparameter settings, and random COCO subsets. The correction study further shows that geometric objective descent and retrieval improvement are distinct outcomes. Together, these results support \papername{} for screening caption quality, tracking modality alignment, and auditing frozen VLM retrieval robustness under low-compute constraints.

\subsubsection*{Declaration of Generative AI Use.}
During the preparation of this work, the authors used OpenAI GPT for language polishing and assistance with data analysis. All AI-assisted text and analytical outputs were reviewed and verified by the authors, who take full responsibility for the content of this publication.

\bibliographystyle{splncs04}
\bibliography{references}

\end{document}